\documentclass[11pt,a4paper]{article}
\usepackage[hyperref]{acl2019}
\usepackage{times}
\usepackage{latexsym}
\usepackage{url}
\usepackage{xspace}
\usepackage{tabularx}

\usepackage{amsmath}
\usepackage{graphicx}
\usepackage{float}
\usepackage{makecell}
\usepackage{algorithm}
\usepackage{multirow}
\usepackage{booktabs}
\usepackage{textcomp}
\usepackage{bbold}
\usepackage[noend]{algpseudocode}
\makeatletter
\def\BState{\State\hskip-\ALG@thistlm}
\makeatother
\usepackage{color}

\newcommand{\hh}[1]{\textcolor{cyan}{\bf\small [#1 --HH]}}

\newcommand{\ignore}[1]{}

\newcommand{\Sref}[1]{\S\ref{#1}}
\newcommand{\fref}[1]{Figure~\ref{#1}}

\newcommand{\tref}[1]{Table~\ref{#1}}

\newcommand{\st}[1]{$\langle \textsl{#1} \rangle$}
\newcommand{\ut}[1]{``\textit{#1}''}

\newenvironment{itemizesquish}{\begin{list}{\labelitemi}{\setlength{\itemsep}{-0.25em}\setlength{\labelwidth}{0.5em}\setlength
{\leftmargin}{\labelwidth}\addtolength{\leftmargin}{\labelsep}}}{\end{list}}

\newcommand\possesivecite[1]{\citeauthor{#1}'s (\citeyear{#1})}
 
\newcolumntype{L}[1]{>{\raggedright\let\newline\\\arraybackslash\hspace{0pt}}m{#1}}
\newcolumntype{C}[1]{>{\centering\let\newline\\\arraybackslash\hspace{0pt}}m{#1}}
\newcolumntype{R}[1]{>{\raggedleft\let\newline\\\arraybackslash\hspace{0pt}}m{#1}}

\aclfinalcopy 

\title{A Dynamic Strategy Coach for Effective Negotiation}

\author{Yiheng Zhou$^{\heartsuit}$ ~ He He$^{\diamondsuit}$ ~ Alan W Black$^{\heartsuit}$ ~ Yulia Tsvetkov$^{\heartsuit}$ \\
$^\heartsuit$Language Technologies Institute, Carnegie Mellon University \\ $^\diamondsuit$Computer Science Department, Stanford University\\
{ \tt $\{$yihengz1, awb, ytsvetko\}@cs.cmu.edu,} { \tt hehe@cs.stanford.edu}}

\date{}

\begin{document}
\maketitle
\begin{abstract}
Negotiation is a complex activity involving strategic reasoning, persuasion, and psychology. 
An average person is often far from an expert in negotiation. 
Our goal is to assist humans to become better negotiators 
through a machine-in-the-loop approach that combines machine's advantage at data-driven decision-making and human's language generation ability.
We consider a bargaining scenario where a seller and a buyer negotiate the price of an item for sale through a text-based dialog.
Our negotiation coach 
monitors messages between them and recommends tactics in real time to the seller to get a better deal
(e.g., ``reject the proposal and propose a price'', ``talk about your personal experience with the product''). 
The best strategy and tactics largely depend on the context (e.g., the current price, the buyer's attitude). 
Therefore, 
we first identify a set of negotiation tactics,
then learn to predict the best strategy and tactics in a given dialog context from
a set of human--human bargaining dialogs.
Evaluation on human--human dialogs shows that 
our coach increases the profits of the seller by almost 60\%.\footnote{The study was approved by the IRB. All sources and data are publicly released at \url{https://github.com/zhouyiheng11/Negotiation-Coach}.}
\end{abstract}

\section{Introduction}

Negotiation is a social activity that 
requires both strategic reasoning and communication skills \citep{thompson2001mind,thompson2010negotiation}.
Even humans require years of training to become a good negotiator.
Past efforts on building automated negotiation agents~\cite{traum2008multi,Heriberto:15,keizer2017evaluating,cao2018emergent, Petukhova2017, Papangelis2017}
has primarily focused on the strategic aspect,
where negotiation is formulated as a sequential decision-making process with a discrete action space,
leaving aside the rhetorical aspect.
Recently, there has been a growing interest in strategic goal-oriented dialog~\cite{he2017symmetric,lewis2017deal,lewis2018hierachical,he2018decouple}
that aims to handle both reasoning and text generation.
While the models are good at learning strategies from human--human dialog and self-play,
there is still a huge gap between machine generated text and human utterances in terms of diversity and coherence~\cite{Li2016, Li2016b}.

\begin{figure}
  \includegraphics[width=5.5cm,height=4cm]{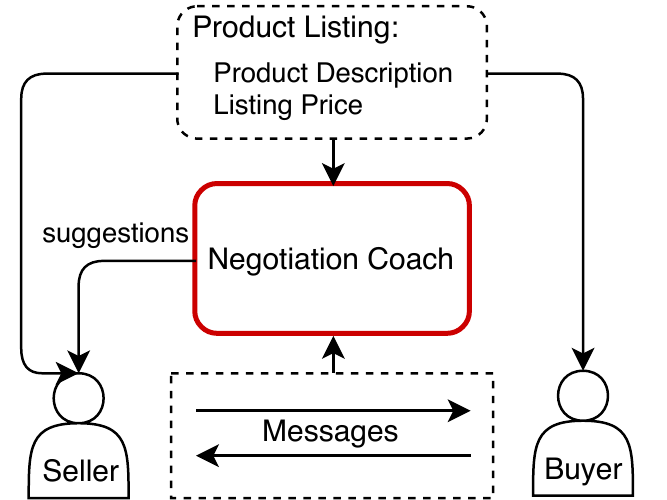}
  \centering
    \caption{
        Our negotiation coach monitors the conversation between the seller and the buyer,
        and provides suggestions of negotiation tactics to the seller in each turn dynamically,  depending on the negotiation scenario, the dialog context, and examples of previous similar dialogs.
    }
   \label{fig:system_architecture}
\end{figure}

In this paper, we introduce a 
machine-in-the-loop
approach~\citep[cf.][]{clark2018creative} that
combines the  
language skills of humans 
and the decision-making skills of machines in negotiation dialogs.
Our \textbf{negotiation coach} assists
users in real time to make good deals in a bargaining scenario between a buyer and a seller.
We focus on
helping the seller to achieve a better deal
by providing suggestions on what to say and how to say it when responding to the buyer at each turn.
As shown in \fref{fig:system_architecture},
during the (human--human) conversation,
our coach analyzes the current dialog history,
and makes both high-level strategic suggestions 
(e.g., \st{propose a price})
and low-level rhetoric suggestions (e.g., \st{use hedge words}).
The seller then relies on these suggestions to formulate their response.

While there exists a huge body of literature on negotiation in behavioral economics~\cite{pruitt2013negotiation,bazerman2000negotiation,William:81,Lax:06,thompson2010negotiation},
these studies typically provide case studies and generic principles such as ``focus on mutual gain''.
Instead of using these abstract, static principles,
we draw insights from prior negotiation literature  and define actionable strategies and tactics conditioned on the negotiation scenario and the dialog context. 
We take a data-driven approach (\Sref{sec:problem}) using human--human negotiation dialogs
collected in a simulated online bargaining 
setting \cite{he2018decouple}.
First, we build detectors to extract negotiation tactics grounded in each turn,
such as product embellishment (\ut{The TV works like a champ!}) and side offers (\ut{I can deliver it to you.}) (\Sref{sec:strategies}).
These turn-level tactics allow us to dynamically predict the tactics used in a next utterance given the dialog context.
To quantify the effectiveness of each tactic,
we further build an outcome predictor to predict the final deal given past tactics sequence extracted from the dialog history (\Sref{sec:success_classifier}). 
At test time, given the dialog history in each turn,
our coach (1) predicts possible tactics in the next turn (\Sref{sec:strategy_predictor}); 
(2) uses the outcome predictor to select tactics that will lead to a good deal; 
(3) retrieves (lexicalized) examples exhibiting the selected tactics and displays them to the seller (\Sref{sec:generating_coaching}). 

To evaluate the effectiveness of our negotiation coach,
we integrate it into \possesivecite{he2018decouple} negotiation dialog chat interface and
deploy the system on Amazon Mechanical Turk (AMT) (\Sref{sec:experiment}).
We compare with two baselines:
the default setting (no coaching)
and the static coaching setting where
a tutorial on effective negotiation strategies and tactics is given to the user upfront.
The results show that our dynamic negotiation coach helps sellers increase profits by 59\% and achieves the highest agreement rate. 


\section{Problem Statement}
\label{sec:problem}

\begin{figure}
\includegraphics[width=0.5\textwidth]{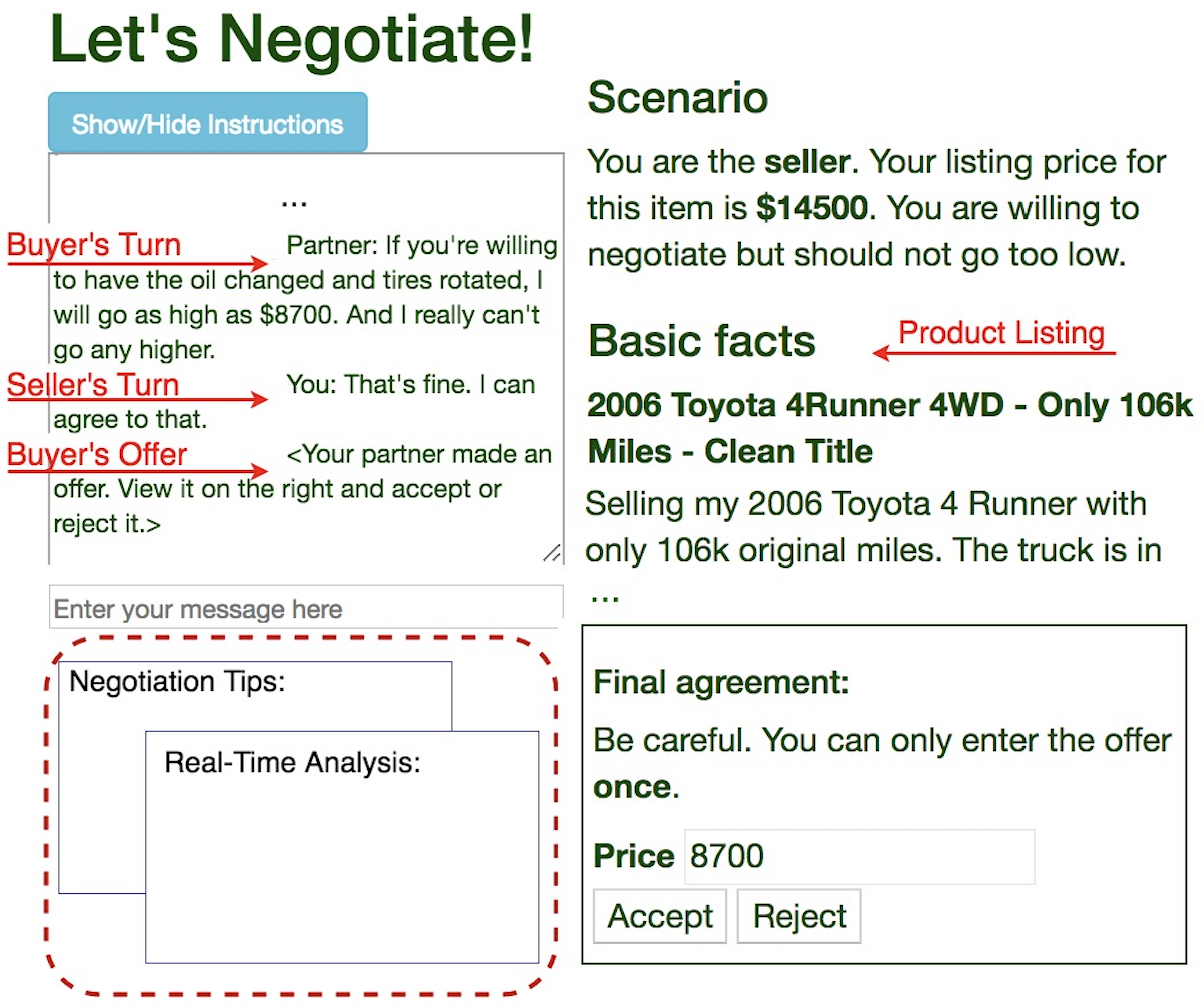}
\caption{Negotiation interface with coaching.} 
\label{fig:negotiation_interface}
\end{figure}

We follow the CraigslistBargain setting of \citet{he2018decouple},
where a buyer and a seller negotiate the price of an item for sale.
The negotiation scenario is based on listings scraped from \url{craigslist.com},
including product description, product photos (if available),
and the listing price.
In addition, the buyer is given a private target price that they aim to achieve.
Two AMT workers are randomly paired to 
play the role of the seller and the buyer.
They negotiate through the chat interface shown in \fref{fig:negotiation_interface} in a strict turn-taking manner.
They are instructed to negotiate hard for a favorable price.
Once an agreement is reached, either party can 
submit the price and the other chooses to accept or reject the deal;
the task is then completed.

Our goal is to help the seller achieve a better deal (i.e. higher final price)
by providing suggestions on how to respond to the buyer during the conversation.
At each seller's turn, the coach takes the negotiation scenario and the current dialog history as input
and predicts the best tactics to use in the next turn to achieve a higher final price.
The seller has the freedom to choose whether to use the recommended tactics.

\section{Approach}
\label{sec:approach}
\begin{figure}
	\includegraphics[width=7.3cm,height=7.5cm]{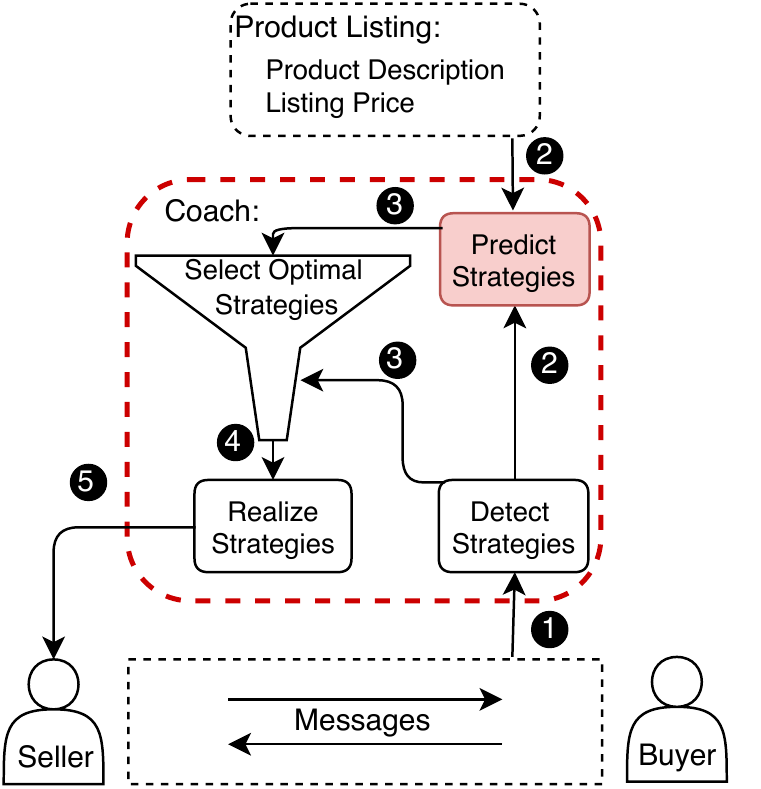}
  	\centering
  	\caption{Negotiation Coach Framework. Numbers indicate the time flow.}
    \label{fig:coaching}
\end{figure}

We define a set of diverse tactics $\mathcal{S}$ from past study on negotiation in behavioral economics,
including both high-level dialog acts (e.g., \st{propose a price}, \st{describe the product}) and low-level lexical features (e.g. \st{use hedge words}).
Given the negotiation scenario and the dialog history,
the coach takes the following steps (\fref{fig:coaching}) to generate suggestions:

\begin{enumerate}
\item The \textbf{tactics detectors} 
map each turn to a set of tactics in $\mathcal{S}$.

\item The \textbf{tactics predictor}
predicts the set of possible tactics in the next turn
given the dialog history. 
For example, if the buyer has proposed a price,
possible tactics include proposing a counter price, agreeing with the price etc.

\item The \textbf{tactics selector} 
takes the candidate tactics from the tactics predictor and
selects those that lead to a better final deal.

\item The \textbf{tactics realizer} 
converts the selected tactics
to instructions and examples in natural language, 
which are then presented to the seller. 
\end{enumerate}

We detail each step in the following sections.

\subsection{Tactics Detectors} 
\label{sec:strategies}
We focus on two broad categories of strategies in behavioral research:
(i) \textbf{integrative}, or win--win, negotiation, 
in which negotiators seek to build 
relationships and reach an agreement benefiting 
both parties; 
and (ii) \textbf{distributive}, or win--lose, negotiation, 
in which negotiators adversarially promote their own interests, 
exert power, bluff, and demand \citep{walton1965behavioral}. 
In practice, effective negotiation often involves 
both types of strategies
\citep[\textit{inter alia}]{William:81,Lax:06,pruitt2013negotiation,de2000influence}. 

Prior work 
typically focuses on conceptual tactics  
(e.g., emphasize mutual interest), 
rather than \emph{actionable} tactics in a specific negotiation scenario 
(e.g., politely decline to lower the price, 
but offer free delivery). 
Therefore, we develop data-driven ways to operationalize and quantify these 
abstract principles.


\begin{table*}[t]
\centering
\footnotesize{
\begin{tabular}{L{2cm}lL{6.5cm}l}
 \toprule
 Principle & Action & Example & Detector \\
 \midrule
 \multicolumn{4}{c}{Integrative strategies} \\
 \midrule
 \multirow{5}{2cm}{Focus on interests, not positions}
 & Describe the product & \ut{The car has leather seats.} & classifier \\
 & Rephrase product description & \ut{45k miles} $\rightarrow$ \ut{less than 50k miles} & classifier \\
 & Embellish the product & \ut{a luxury car with attractive leather seats} & classifier \\
 & Address buyer's concerns & \ut{I've just taken it to maintainence.} & classifier \\
 & Communicate your interests & \ut{I'd like to sell it asap.} & classifier \\
 \midrule
 \multirow{3}{2cm}{Invent options for mutual gain}
 & Propose a price & \ut{How about \$9k?} & classifier \\
 & Do not propose first & n/a & rule \\
 & Negotiate side offers & \ut{I can deliver it for you} & rule \\ 
 & Use hedges & \ut{I could come down a bit.} & rule \\
 \midrule
 \multirow{3}{2cm}{Build trust}
 & Communicate politely & greetings, gratitude, apology, \ut{please} & rule \\
 & Build rapport & \ut{My kid really liked this bike, but he outgrew it.} & rule \\
 & Talk informally & \ut{Absolutely, ask away!} & rule \\
 \midrule
 \multicolumn{4}{c}{Distributive strategies} \\
 \midrule
 \multirow{3}{2cm}{Insist on your position}
 & Show dominance & \ut{The absolute highest I can do is 640.0.} & rule \\
 & Express negative sentiment & \ut{Sadly I simply cannot go under 500 dollars.} & rule \\
 & Use certainty words & \ut{It has always had a screen protector} & rule \\
 \bottomrule
\end{tabular}
}
\caption{
    Actionable tactics designed based on negotiation principles.
    Some of them are detected by learning classifiers on annotated data,
    and the rest are detected using pattern matching.
}
\label{tab:strategy}
\end{table*}

In \tref{tab:strategy}, we list our actionable tactics  motivated by various negotiation principles. 
To detect these tactics from turns,
we use a mix of learned classifiers\footnote{We use $\ell_2$-regularized Logistic Regression classifiers.} for turn-level tactics (e.g., propose prices)
and regular expression rules for lexical tactics (e.g., use polite words).
To create the training set for learning tactic   predictors, we randomly selected 200 dialogs and annotated them with tactics.\footnote{Each turn can be labeled with multiple tactics.}
The detectors use the following features: 
(1) the number of words overlapping with the product description; 
(2) the METEOR score \citep{denkowski2014meteor} of the turn given the product description as reference; 
(3) the cosine distance between the turn embedding and the product description embedding.\footnote{Sentence embeddings were calculated as the mean of the word embeddings. We used pre-trained word2vec embeddings \cite{Mikolov:13}.} 
For ``Address buyer's concerns'', we additionally include lexical features indicating a question (e.g.,\ut{why}, \ut{how}, \ut{does}) from the immediate previous buyer's turns. 
\tref{tab:sd1-4} summarizes the number pf training examples and prediction accuracies for each learned classifier. 
For lexical tactics, we have the following 
rules:
\begin{itemizesquish}
\item \st{Do not propose first}\\ Waiting for the buyer's proposal allows the seller to
better estimate 
the buyer's target. The detector simply keeps track of who proposes a price first by detecting \st{propose a price}. 
\item \st{Negotiate side offers}\\
The seller sometimes negotiates side offers, e.g., 
offering a free gift card or free delivery. 
To detect this strategy, we match the turn against a set of phrases, 
e.g., \ut{throw in}, \ut{throwing in}, \ut{deliver}, 
\ut{delivery}, \ut{pick up}, \ut{pick it up}, \ut{in cash}.

\item \st{Use factive verbs} \\defined in \cite{Joan:75} (e.g. \textit{know});
\item \st{Use hedge words} \\defined in \cite{ken:05} (e.g. \textit{could, would});
\item \st{Use certainty words}\\ defined in the LIWC dictionary~\cite{tausczik2010psychological}.

\item \st{Communicate politely} \\
We include several politeness-related negotiation 
tactics that were identified by \citet{Danescu:13} 
as most informative features. They include: gratitude, greetings , apology, ``please'' in the beginning of a turn, 
``please'' later on. Keywords matching is used to detect these tactics. 

\item \st{Build rapport}\\
Deepening self-disclosure, e.g., ``My kid really liked this bike, but he outgrew it'', is one strategy for building rapport. We implemented three tactics detectors to identify self-disclosure. First, we count first-person pronouns \cite{Valerian:87,Adam:01}. 
Second, we count mentions of family members and friends, respectively \cite{wang:16}. 
It is done by matching lexicons from \textit{family} and \textit{friend} categories in LIWC.

\item \st{Talk informally} \\ It is detected by matching 
the keywords in the \textit{informal language} category in LIWC. 

\item \st{Show dominance} \\ To detect stubbornness \cite{Chenhao:16}, we measure the average dominance score of all the words from the \possesivecite{Amy:13}'s dominance ratings of 14,000 words.

\item \st{Express negative sentiment}\\ 
We measure both positive and negative sentiment 
by counting words from \textit{positive} and \textit{negative} categories in LIWC.
\end{itemizesquish}

\begin{table}[!h]
\centering
\begin{tabular}{L{5cm} r r } 
 \toprule
  Strategy & \# Ex & Acc \\ 
 \midrule
Describe the product & 228 & 0.88 \\
Rephrase product description & 136 & 0.74 \\
Embellish the product & 200 & 0.70 \\
Address buyer's concerns & 192 & 0.95 \\
Propose a price & 290 & 0.88 \\

 \bottomrule
\end{tabular}
\caption{Number of turns annotated (\# Ex) and prediction accuracies (Acc) by 5-fold cross validation for learned strategy predictors.
Our classifiers achieve high accuracy on all tactics.}
\label{tab:sd1-4}
\end{table}

\section{Tactics Predictor}
\label{sec:strategy_predictor}
Armed with a set of negotiation tactics from the dataset, 
the tactics predictor monitors a negotiation conversation and, at each turn, 
predicts the seller's next move (e.g., \st{propose a price} or \st{express negative sentiment}) 
given the current dialog context. 

Let $u_1,...,u_t$ denote a sequence of turns, $d$ be a product category,
and $o_t$ be a set of tactics occurred in turn $u_t$.
At the ($t+1$)-th turn in a dialog, given the current dialog context $u_{1:t}$ and $d$,
we want to predict what tactics to use in the response,
i.e.~$o_{t+1}$.

The dialog context is represented by embedding 
the turns, tactics extracted from the turns (\Sref{sec:strategies}),
and the product being discussed.
The set of tactics $o$ is
a binary vector, where each dimension corresponds to the existence of a certain tactic.


\paragraph{Embedding the turns}
Embedding of the turns is computed using a standard LSTM encoder over concatenated sequences of words $x_i$ in each turn:

\begin{align*}
h^u_i = 
\text{LSTM}^u\left(h^u_{i-1}, E^w(x_{i-1})\right),
\end{align*}
where $E^w$ is the word embedding to be learned.

\paragraph{Embedding the tactics}
By using the tactics detectors from ~\Sref{sec:strategies}, we extract a sequence of tactics $\{m_i\}$ for each turn $u$ in the order of their occurrences from left to right.
For example, \ut{Hi there, I've been using this phone for 2 years and it never had any problem.}
is mapped to
``\st{greetings} \st{use certainty words}''.
Given turns $u_{1:t}$, we concatenate their tactics in order to form a single sequence,
which is embedded by an LSTM:
\begin{align*}
h^s_i = \text{LSTM}^s\left(h^s_{i-1}, [E^o(m_{i-1}); b_{i-1}])\right),
\end{align*}
where $E^o$ is the one-hot embedding and $b$ is a binary vector encoding tactics that are not specific to a particular word $x_i$ but occur at the turn level (e.g. \st{describe the product}).



\paragraph{Embedding the product}
Different products often induce different expressions and possibly different tactics; for example, renting an apartment often has conversation about a parking lot while selling a phone does not.
Thus we also include the product embedding, $E^p$ to encode the product category $d$, including \textit{car, house, electronics, bike, furniture, and phone}. 

The output set of tactics $o_{t+1}$ is a 24-dimensional \footnote{Table \ref{tab:strategy} contains only 15 tactics because some tactics consist of multiple sub-tactics. For example, \st{build rapport} includes two sub-tactics: \st{mention family members} and \st{mention friends}.} binary vector,
where each dimension represents whether a certain tactic occurred in $u_{t+1}$.
Given the context embedding, we compute the probability of the $j$-th tactic occurring in $u_{t+1}$ by
\begin{align*}
p(o_{t+1,j} | u_{1:t}, d) = \sigma(W_j[h^s_{t};h^u_{t};E^p(d)] + b_j),
\end{align*}
where $h^s_{t}$ and $h^u_{t}$ are final hidden states of the tactics encoder and the utterance encoder respectively,
and $W_j$ and $b_j$ are learnable parameters.
We train the predictor by maximizing the log likelihood of tactics.

\subsection{Evaluation of the Tactics Predictor}
We evaluate the effect of different embeddings on predicting next tactics. We split our data into train, held-out development (20\%) and test (20\%) data. We then remove incomplete 
negotiation dialogs 
(e.g. when the chat got disconnected in the middle).
Data sizes are 1,740, 647, and 527 dialogs for train, development and test data respectively. 
We initialize word embeddings with pre-trained word2vec embeddings.
The LSTMs have 100 hidden units.
We apply a dropout rate of~0.5 and train for~15 epochs with~SGD.


Given the output probabilities $p(o_j)$,
we need a list of thresholds $\gamma$ to convert it into a binary vector,
such that $o_j = \mathbb{1}(o_j > \gamma_j)$.
We choose $\gamma$ by maximizing the F1 score of the corresponding strategy on the development set.
Specifically, for each strategy, we iterate through all threshold values $[0,1]$ with a step size of 0.001
and select the one that produces the highest F1 score.

We conduct an ablation study and calculate micro and macro F1 scores. 
As shown in \tref{tab:ablation}, we achieve the best result 
when combining all components. 
\begin{table}[!h]
\centering
\begin{tabular}{lcc} 
 \toprule
  Components & Macro F1 & Micro F1\\
 \midrule
 Turn Embedding & 0.382 & 0.536\\ 
 $+$Product Embedding& 0.384 & 0.539\\ 
 ~ $+$Tactics Embedding & \textbf{0.397} & \textbf{0.592}\\ 
 \bottomrule
\end{tabular}
\caption{Effectiveness of turn, product, and tactics embeddings in predicting the next move.}
\label{tab:ablation}
\end{table}

\section{Tactics Selector}
\label{sec:success_classifier}
The tactics predictor outputs a set of tactics $o_{t+1}$, which can be non-optimal because we only model human behaviors. 
Now, we implement a \textit{tactics selector} that selects optimal tactics from $o_{t+1}$ under the current dialog context. 
The major component of the selector is a \textit{negotiation outcome classifier}. 
This is a supervised classifier  
that predicts a binary outcome of whether the negotiation 
will be successful from the seller's standpoint. 
We next describe the classifier and its evaluation. 

Given negotiation tactics and  word and phrase choices 
used by both parties in the previous turns, 
we train a $\ell_2$-regularized Logistic Regression classifier to 
predict the negotiation's outcome. 
The outcome is defined as \textit{sale-to-list} ratio $r$, which   
is a standard valuation ratio in sales, 
corresponding to the ratio between 
the final sale price (i.e., what a buyer pays for the product) 
and the price originally listed by the seller, 
optionally smoothed by the buyer's target price (Eq.~\ref{eq1}). 
If the agreed price is between the listed price and the 
buyer's budget, then $0 \leq r \leq 1$. 
If the agreed price is greater than the listed price, then $r > 1$. 
If the agreed price is less than the buyer's budget, then $r < 0$.  
We define a negotiation as successful if 
its sale-to-list ratio is in the top 22\% of all 
negotiations in our training data; negative examples comprise the bottom 22\%.\footnote{The thresholds were set empirically during an early experimentation with the training data.} 
\begin{equation}
\label{eq1}
r = \dfrac{\text{sale price} - \text{buyer target price}}{\text{listed price} - \text{buyer target price}}
\end{equation}

The features are the counts of each negotiation tactic from \Sref{sec:strategies}, separately for the seller and the buyer. 
A typical negotiation often involves a smalltalk in 
the beginning of the conversation. 
Therefore, we split a negotiation into two stages: 
the $1^{st}$ stage consists of turns that happen before the first price was proposed, 
and the $2^{nd}$ stage includes the rest. 
We count each tactic separately for the two stages. 

Lastly, we apply the classifier to select tactics  
that will make the negotiation more successful. 
For each tactic in $o_{t+1}$, 
we assume that the seller will use it next by modifying the corresponding input feature in the classifier, which outputs the probability of a successful negotiation outcome for the seller.
If the modification results in a more successful negotiation, 
we select the tactic. 
For example, if incrementing the input feature of 
\st{describe the product} $\in o_{t+1}$ increases the 
probability outputted by the outcome classifier, we select \st{describe the product}.

\subsection{Evaluation of the Outcome Classifier}
\label{implementation_class} 
The accuracy on test data from \tref{table:data} is given in \tref{table:succ_predictor}. 
We also evaluate a baseline with shallow lexical features (1-, 2-, 3-grams). 

\begin{table}[!h]
\centering
\begin{tabular}{lrcc} 
 \toprule
  & Total & Successful & Unsuccessful\\ 
 \midrule
Training & 1,740 & 872& 868\\
 Dev & 647& 316& 331\\
 Test & 527& 259& 268\\ 
 \bottomrule
\end{tabular}
\caption{Statistics of dialogs, split by successful/unsuccessful negotiations from the seller's standpoint.}
\label{table:data}
\end{table}
\begin{table}[!h]
\centering
\begin{tabular}{l c} 
\toprule
Features & Accuracy\\
\midrule
Shallow features & 0.60\\
Strategy-motivated features & \textbf{0.83}\\
\bottomrule
\end{tabular}
\caption{Test accuracy of the outcome classifier with different feature groups}
\label{table:succ_predictor}
\end{table}


One contribution of this work is that we not only 
present abstract tactics recommendations (e.g. \st{propose a price}), 
but also propose lexical tactics and examples from successful negotiations (e.g. ``Try to use the word \textit{would} like in this sentence: \ldots''). 
Table \ref{table:ablation_succ_predictor} shows that removal of the lexical tactics drops the accuracy by 11\%, which is similar to the removal of abstract negotiation tactics. 
We also find that it is important to separate 
features in the two stages (before/after the first offer). 
The $1^{st}$ stage has weaker influence on the success, while the removal of features in $2^{nd}$ stage makes the accuracy drop by 24\%. Features from both stages contribute to the final score. 
\begin{table}[!h]
\centering
\small
\begin{tabular}{l c} 
\toprule
Removed Features & $\Delta$ Accuracy\\
\midrule
Abstract strategies & -0.12\\
Lexical strategies & -0.11\\
Features from the $1^{st}$ stage & -0.02\\
Features from the $2^{nd}$ stage & -0.24\\
\bottomrule
\end{tabular}
\caption{Ablation of each subset features shows that lexical tactics are equally important as higher-level abstract tactics and both stages contribute to the final score.}
\label{table:ablation_succ_predictor}
\end{table}

We list seller's top weighted negotiation tactics for both stages in \tref{table:feature_analysis}. 
\st{propose a price} has the highest weight, which is expected because giving an offer is a fundamental action of negotiation.\footnote{The reason that \st{propose a price} has zero weights in the $1^{st}$ stage is that the $1^{st}$ stage is defined to be the conversations before any proposal is given.} Following that, the negative weight of \st{do not propose first} indicates that seller should wait for buyer to propose the first price. It is probably because the seller can have a better estimation of the buyer's target price. The second most weighted strategy in the $2^{nd}$ stage is \st{negotiate side offers}, which emphasizes the importance of exploring side offers to increase mutual gain. Moreover, building rapport can help develop trust and help get a better deal, which is supported by the positive weights of \st{build rapport}.

Interestingly, some strategies are effective only in one stage, but not in the other (the strategies with an opposite sign). For example, \st{talk informally} is more preferable in the $1^{st}$ stage where people exchange information and establish relationship, while trying to further reduce social distance in the $2^{nd}$ can damage seller's profit. Another example is that \st{express negative sentiment} is not advised in the $1^{st}$ stage but has a high positive weight in the $2^{nd}$ stage. Overall these make sense: to get to a better deal the seller should be friendly in the $1^{st}$ stage, but firm, less nice, and more assertive in the $2^{nd}$, when  negotiating the price. 

\begin{table}[H]
\centering
\small
\begin{tabular}{lrr} 
 \toprule
  Features & \makecell{$1^{st}$ stage \\ Weights} & \makecell{$2^{nd}$ stage \\ Weights}\\
\midrule
\st{propose a price} & 0.0 & 2.28\\
\st{do not propose first} &-0.62 & -0.62\\
\st{negotiate side offers} & -0.27 & 1.11\\
\st{build rapport} &0.08&0.26\\
\st{talk informally} & 0.39 & -0.39\\ 
\st{express negative sentiment} &-0.05&0.61\\
\bottomrule
\end{tabular}
\caption{The table shows the weights of seller's top weighted negotiation tactics in both stages. Positive weight means the feature is positively correlated with the success of a negotiation.
}
\label{table:feature_analysis}
\end{table}



\section{Giving Actionable Recommendations}
\label{sec:generating_coaching}
Finally, given the selected tactics, 
the coach provides suggestions in natural language to the seller. 
We manually constructed a set of natural language suggestions 
that correspond to all possible combinations of strategies. For example, if the given tactics are \{\st{describe the product}; \st{propose a price}; \st{express negative sentiment}\}, then the corresponding suggestion is "\textit{Reject the buyer's offer and propose a new price, provide a reason for the price using content from the Product Description}. 

As discussed above, we also retrieved examples of some tactics. 
For instance, \st{use hedges} is not a clear suggestion to most people. 
To retrieve best examples of \st{use hedges}, 
from all the turns that contain \st{use hedges} in the training data, 
we choose the one that has a most similar set of tactics 
to the set of tactics in the current dialog. 

\section{End-to-End Coaching Evaluation}
\label{sec:experiment}
We evaluate our negotiation coach by incorporating into mock negotiations on AMT. We compare the outcomes of negotiations using our coach, using a static coach, and using no coach.

\subsection{Setup and Data}
We modified the same interface that was used for collecting data in \Sref{sec:problem} for the experiments. Moreover, we created 6 test scenarios for the experiments and each scenario was chosen randomly for each negotiation task. 
\begin{itemizesquish}
\item \textbf{No coaching} For our baseline condition, we leave the interface unchanged and collect human--human chats without any interventions, as described in \Sref{sec:problem}.
\item \textbf{Static coaching} We add a box called "Negotiation Tips", which is shown in a red dashed square in Figure \ref{fig:negotiation_interface}. At the beginning of each negotiation, we ask sellers to read the tips. The tips encourage the seller to use a subset of negotiation tactics in \Sref{sec:strategies}:
\begin{itemizesquish}
\item[--] Use product description to negotiate the price.
\item[--] Do not propose price before the buyer does.
\item[--] You can propose a higher price but also give the buyer a gift card.
\item[--] You can mention your family when rejecting buyer's unreasonable offer, e.g., my wife/husband won't let me go that low.
\end{itemizesquish}
Only a subset of tactics was used: the most important and most clear tactics that fit in the recommendation window. 

\item \textbf{Dynamic coaching} We replace "Negotiation Tips" with "Real-Time Analysis" box as shown in Figure \ref{fig:negotiation_interface}. When it is the seller's turn to reply, the negotiation coach takes the current dialog context and updates the "Real-Time Analysis" box with contextualized suggestions.
\end{itemizesquish}



We published three batches of assignments on AMT for three coaching conditions and only allow workers with greater than or equal to 95\% approval rate, location in US, UK and Canada to do our assignments. Before negotiation starts, each participant is randomly paired with another participant and appointed to either seller or buyer. During negotiation, seller and buyer take turns to send text messages through an input box. The negotiation ends when one side accepts or rejects the final offer submitted by the other side, or either side disconnects. 

We collected 482 dialogs over 3 days. We removed negotiations with 4 turns or less.\footnote{Sometimes sellers offered a price much lower than the listing price in order to complete the task quickly.} We further remove negotiations where the seller followed our suggested tactics less than 20\% of the time (only 6 dialogs are removed). Our final dataset consists of 300 dialogs, 100 per each coaching condition\footnote{We randomly sampled 100 dialogs from 108 for no-coaching} In the 300 final dialogs, 594 out of 600 workers were unique, only 6 workers participated in negotiations more than once.

\subsection{Result}

We use two metrics to evaluate each coaching condition: average \textit{sale-to-list} ratio (defined in \Sref{sec:success_classifier}) and task completion rate (\%Completion), the percentage of negotiations that have agreements. Moreover, to measure increase in profits ($\Delta$\%Profit), we calculate the percentage increase in sale-to-list ratio comparing to no coaching baseline. The result is in Table \ref{table:result}. Dynamic coaching achieves significantly higher sale-to-list ratio than the other coaching conditions, and it also has the highest task completion rate. Comparing with no coaching baseline, our negotiation coach helps the seller increase profits by 59\%. 

\begin{table}[!h]
\centering
\small
\begin{tabular}{lrrr} 
\toprule
 & \makecell{No \\Coaching} & \makecell{Static\\ Coaching} & \makecell{Dynamic\\ Coaching}\\
 \midrule
Sale-to-List & 0.22 & 0.19 & \textbf{0.35}\\
$\Delta$\%Profit & - & -13.6\%& \textbf{+59.0\%}\\
\midrule
\%Completion& 66\% & 51\% & \textbf{83\%}\\
 \bottomrule
\end{tabular}
\caption{Evaluation of three coaching models. Improvements are statistically significant ($p< 0.05$).}
\label{table:result}
\end{table}

\subsection{Analysis}
Here, we first explore the reasons for effectiveness of our dynamic coach and then study why static coaching is least useful.

\ignore{
\begin{table}[!htbp]
\centering
\small
\begin{tabular}{llll}
 \toprule
 &\makecell{No \\Coaching} & \makecell{Static\\ Coaching} & \makecell{Dynamic\\ Coaching}\\
\midrule
$\textsc{SD}_{7}$ &1.32$\pm0.48$&1.08$\pm0.50$&\textbf{1.93}$\pm0.45$\\
\%ReProp & 0.61$\pm0.29$ &  0.69$\pm0.31$ &  \textbf{0.96}$\pm0.11$\\
\%BadStrat & 0.33$\pm0.29$ &  0.38$\pm0.3$ &  \textbf{0.26}$\pm0.21$\\
\#UniStrat&7.23$\pm3.5$ & 7.35$\pm3.3$ &\textbf{9.22}$\pm2.2$\\
\#Strat & 12.48$\pm7.9$ &  12.26$\pm8.5$ &  \textbf{16.09}$\pm6.4$\\
\bottomrule
\end{tabular}
\caption{Other measurements of the collected data show the reasons that dynamic coaching achieves the best result.}
\label{table:dynamic_evaluation}
\end{table}
}

\paragraph{Why is dynamic coaching better?} 
Manual analysis reveals that our coach encourages sellers to be more assertive while negotiating prices, whereas sellers without our coach give in more easily.\footnote{For an example, refer to Table \ref{table:case_study} in the Appendix; compare lines 24, 26, 28 (our system) against lines 4, 6, 14, 16. } 
We measure \emph{assertiveness} with the average number of proposals made by sellers \st{propose a price}: 
sellers with dynamic coaching propose more often (1.93, compared to 1.32 and 1.08 for no coaching and static coaching respectively). The average number of turns is 8; the measured assertiveness of our coach (1.93) shows that we do not always suggest the seller to reject the buyer's proposal.

Intuitively, an assertive strategy could annoy the buyer and make them leave without completing the negotiation. But, negotiations using our coach have the highest task completion rate. 
This is likely because in addition to encouraging assertiveness, our coach suggests additional actionable tactics to make the proposal more acceptable to the buyer.  
We find that 96\% of the time, sellers with dynamic coaching use additional  strategies when proposing a price, as compared to 69\% in static coaching and 61\% with no coaching. 
For example, our coach suggests the seller negotiate side offers and use linguistic hedges, which can mitigate the assertiveness of the request. 
On the other hand, in no coaching settings, sellers often propose a price without using other tactics. 
Lastly, the seller often uses almost the same words as shown in the examples retrieved by our suggestions generator in \Sref{sec:generating_coaching}. This is probably because sellers find it easier to copy the retrieved example than come up with their own.


The effectiveness of dynamic coaching could in large part be attributed to the \emph{tactics selector} that selects optimal tactics under the current dialog context, but sellers might still use non-optimal tactics even if they are not suggested. To observe the effect of this selecting, we compute the average percentage of \emph{non-optimally} applied tactics.  
Dynamic coaching has the lowest rate (26\%), as compared to no coaching (33\%) and static coaching (38\%). 
Moreover, we find that sellers with dynamic coaching often have different chatting styles for exchanging information ($1^{st}$ stage) and negotiating price, while sellers without our coach often use the same style. 
For example, we show several turns from two dialogs ($\text{D}_1$, $\text{D}_2$) for dynamic and no coaching, respectively. 
In the $1^{st}$ stage, our coach suggests sellers to \st{talk informally} with positive sentiment:

\begin{itemizesquish}
\item $\text{D}_1$ with dynamic coaching:\\
\textbf{Buyer:} "\textit{I'd like to buy the truck.}"
\\\textbf{Seller:} "\textit{well that's great to hear! Only 106k miles on it and it runs amazingly. I've got a lot on my plate right now lol so I priced this lower to move it quickly}".
\item $\text{D}_2$ with no coaching:\\
\textbf{Buyer:} "\textit{I am interested in this truck but I have a few questions.}"
\\\textbf{Seller:} "\textit{Absolutely, ask away!}"
\end{itemizesquish}
The sellers in both dialogs chat in a positive and informal way. However, when negotiating the price, our coach chooses not to select \st{talk informally}, but instead suggests formality and politeness, and \st{express negative sentiment} when rejecting buyer's proposal:
\begin{itemizesquish}
\item  $\text{D}_1$ with dynamic coaching:\\
\textbf{Buyer:} "\textit{Would you be willing to take 10k?}"
\\\textbf{Seller:} "\textit{That's a lot lower than I was hoping. what I could do, is if you wanted to come see it I could knock off \$1500 if you wanted to buy.}".
\item  $\text{D}_2$ with no coaching:\\
\textbf{Buyer:} "\textit{I'm looking for around 10,000.}"
\\\textbf{Seller:} "\textit{Oh no. Lol. That's way too low!}"
\end{itemizesquish}

While the seller with our coach changes style, the seller with no coaching stays the same. We attribute this to the tactics selector.
We also find that dynamic coaching leads to a larger quantity and a richer diversity of tactics.

Lastly, we focus on diversity: we show that our coach almost always gives recommendations at each turn and does not recommend the same tactics in each dialog. Specifically, we measure how often our coach gives no suggestions and find out that only 1.8\% of the time our coach recommends nothing (9 out of 487 sellers' turns).  Then, we calculate how often our coach gives the same tactics within each dialog and find out that only 10\% of the time our coach gives the same suggestions (49 out of 487 sellers' turns).

\ignore{
\paragraph{Effective and ineffective suggestions.} Next, we investigate the effectiveness of each suggested strategy, because sellers may use some common strategies even if they are not suggested. For example, \st{address buyer's concerns} is a natural reaction when the buyer asks a question. 
We start by inputting each dialog with no coaching to our negotiation coach and getting a set of suggested strategies $o_t$ at each turn $u_t$. Intuitively, if a strategy used in $u_t$ also appears in $o_t$, then suggesting this strategy is not effective. Therefore, for each strategy, we measure how many times it appears in both $u_t$ and $o_t$ for all possible $t$, smoothed by the frequency of this strategy in negotiations with dynamic coaching. The top three most ineffective (commonly-used) strategies are \st{address buyer's concern} as expected, \st{talk about yourself}, and \st{using words with certainty}. The top three most effective strategies: \st{describe the product}, \st{greet}, and \st{express negative sentiments} when negotiating prices.
}

\paragraph{Why is static coaching even worse than no coaching?} Surprisingly, static coaching has even lower scores in both metrics than no coaching does. Two possibilities are considered. One is that reading negotiation tips can limit seller's ability to think of other tactics, but we find that static and dynamic coaching use similar number of unique tactics. 
Then, we explore the second possibility: it is worse to use the tactics in the tips under non-optimal context. Therefore, we measure the average percentage of \emph{non-optimally} applied strategies, but only consider the tactics mentioned in the tips. 
The result shows that static coaching uses non-optimal tactics 51\% of the time, compared to 46\% and 38\% for no coaching and dynamic coaching, respectively.

\ignore{
\section{Related Work}
\hh{
1. negotiation dialogs. Lewis et al, He et al. hard to learn both good strategy and good language realization.
Big gap between bot and human.
2. negotiation policy. Traum et al, Cao et al, only focus on action (e.g. price).
3. any work on education that uses coaching / tutoring?
Emphasize the lingustics cues in our strategies.
}

In the field of natural language processing, \cite{lewis2017deal} implemented a chat bot that negotiates divisions of a collection of three kinds of items between two agents. 
There are at least two reasons. The first reason is that their tasks do not require complicated use of language. Second reason is that there are no available negotiation corpus that allow them to learn linguistic insights from.

There has been a growing interest in studying automated negotiation agents and strategies \cite{Tsung:16} \cite{he2017symmetric} \cite{lewis2017deal} \cite{Heriberto:15} \cite{Avi:2014} \cite{Akiyuki:16}. However, most of them ignore rhetorical aspect of negotiation agents. In particular, \cite{Avi:2014} implemented a list of dialog acts as intermediates between natural language and their system, which is similar to our approach, but they have a very limited number of dialog acts and none of them are rhetorical. Moreover, they designed a rule-based system that does not integrate any studies on negotiation theories.
}

\section{Conclusion}

This paper presents a dynamic negotiation coach that can make measurably good recommendations to sellers that can increase their profits.  
It benefits from grounding in strategies and tactics within the negotiation literature and uses natural language processing and machine learning techniques to identify and score the tactics' likelihood of being successful.  
We have tested this coach on human--human negotiations and shown that our techniques can substantially  increase the profit of negotiators who follow our coach's recommendations. 


A key contribution of this study is 
a new task and a framework of an automated coach-in-the-loop 
that provides on-the-fly autocomplete suggestions 
to the negotiating parties. 
This framework can seamlessly be integrated in 
goal-oriented negotiation dialog systems \citep{lewis2017deal,he2018decouple}, 
and it also has stand-alone educational and commercial values. 
For example, our coach can provide language and strategy guidance 
and help improve negotiation skills of non-expert negotiators. 
In commercial settings, it has a clear use case of 
assisting humans in sales and in customer service.  
An additional important contribution lies in aggregating 
negotiation strategies from economics and behavioral research, 
and proposing novel ways to operationalize the strategies using 
linguistic knowledge and resources. 

\section*{Acknowledgments}
 We gratefully thank Silvia Saccardo, Anjalie Field, Sachin Kumar, Emily Ahn, Gayatri Bhat, and Aldrian Muis for their helpful feedback and suggestions. 
 
\bibliography{acl2019}
\bibliographystyle{acl_natbib}

\section{Appendix}
\label{sec:appendix}
\begin{table*}[h]
\newcolumntype{b}{X}
\newcolumntype{s}{>{\hsize=.01\hsize}X}
\newcolumntype{m}{>{\hsize=.05\hsize}X}
\centering
\small
\begin{tabularx}{\linewidth}{XXX}
\toprule
Product Listing:\\
\midrule
\textbf{Listing Price:} 14500\\
\textbf{Buyer's Target Price:} 8700\\
\textbf{Title:} "2006 Toyota 4Runner 4WD - Only 106k Miles - Clean Title"\\
\textbf{Product Description}:\\
"Selling my 2006 Toyota 4 Runner with only 106k original miles. The truck is in great condition with no mechanical flaws whatsoever and a clean accident history. Got new tires about 3,000 miles ago. Always has the oil changed on time (due in about 1k). Just got a thorough cleaning inside and a wash and wax outside (still wet in the photos). This truck has never been offroad, but the 4WD is working perfectly from the few times we've been up to Tahoe in it. However, it's a 10+ year old truck that's been driven, not babied and garaged all the time. It's got some scratches, paint is not perfect, but zero body damage."\\
\bottomrule
\end{tabularx}
\begin{minipage}{.5\linewidth}
\begin{tabularx}{\linewidth}{X|}
\toprule
No Coaching:\\
Seller: S  Buyer: B\\
\end{tabularx}
\begin{tabularx}{\linewidth}{ssb|}
\midrule
\textit{1.} & B: & I just saw your ad for the 4Runner, can you send more picture of the scratches?\\
\textit{2.} & S: & I don't have pictures of the scratches but I can assure you it's minor\\
\textit{3.} & B: & I might be interested,but all I can offer is \$7500\\
\textit{4.} & S: & That is very low. Can I agree to 11000?\\
\textit{5.} & B: & That is too high for me, I mean it is 10 years old with over 100,000 miles. I can possible come up to \$8,000\\
\textit{6.} & S: & I can agree to 9,000 and make sure it's had a oil change and tire rotation before you pick it up.\\
\textit{7.} & B: & If you're willing to have the oil changed and tires rotated, I will go as high as \$8700. And I really can't go any higher.\\
\textit{8.} & S: & That's fine. I can agree to that.\\
\textit{9.} & B: & Thanks, I'll be right over to pick it up.\\
\bottomrule
\end{tabularx}
\end{minipage}%
\begin{minipage}{.5\linewidth}
\begin{tabularx}{\linewidth}{X}
\toprule
Static Coaching:\\
Seller: S  Buyer: B\\
\end{tabularx}
\begin{tabularx}{\linewidth}{ssb}
\midrule
\textit{10.} & S: &i would sell it for 1400\\
\textit{11.} & B: &you got a deal. 1400 it is\\
\textit{12.} & S: &sorry meant 14000\\
\textit{13.} & B: &Oh c'mon. Now you got my hopes up. lol\\
\textit{14.} & S: &i can go low around 12000\\
\textit{15.} & B: &I'm looking at the blue book right now and that still seems a bit high.\\
\textit{16.} & S: &well the lowest i can go is 10000\\
\textit{17.} & B: &You mention scratches. Lets be real. How bad are we talking?\\
\textit{18.} & S: &its 10 yrs old it has some scratches but has zero body damage.\\
\textit{19.} & B: &Ok. Without seeing any more photos, 9000 is the best I can do. \\
\textit{20.} & S: &deal\\
\textit{21.} & B: &deal. Thanks!\\
\bottomrule
\end{tabularx}
\end{minipage}
\begin{tabularx}{\linewidth}{X}
\toprule
Dynamic Coaching: \\
Seller: S  Buyer: B\\
\end{tabularx}
\begin{tabularx}{\linewidth}{smb}
\hline
\textit{22.} & S: &Would you be interested in buying my 4Runner?\\
\textit{23.} & B: &Yes, I am possibly. interested Given that is is over 10 years old, all I can offer now is 8000.\\
\textit{24.} &Coach: &\textit{Reject Buyer's Offer and Propose a New Price, Reason the Price with Content in the Product Description}\\
  &&\textit{Try to Use the Word "Would" Like This: "Sorry, I would really need \$100 for it."}\\ 
&&\textit{...}\\ 
& S: &I'm sorry, but I \textbf{would} really need \$12,000 for it. \textbf{It's in great condition!}\\
\textit{25.} & B: &Well given the mileage on the car it is about to be due for some major service which can be expensive. How about 9000?\\
\textit{26.} & Coach: &\textit{Reject Buyer's Offer and Propose a New Price}\\
&&\textit{Try to Use the Word "Could" Like This: "I could come down to \$3."}\\
&&\textit{You Can Also Give the Buyer Something for Free (Gift Card) to Argue for Higher Price.}\\
&&\textit{...}\\
& S: &I \textbf{could} come down to \$11,000 and \textbf{include the roof rack as a bonus}.\\
\textit{27.} & B: &Add in a full tank of gas and a cold beer and I could do 10500.\\
\textit{28.}& Coach: &\textit{Reject Buyer's Offer and Propose a New Price}\\
&&\textit{Talk About Your Personal Experience With The Product}\\
&&\textit{Try to Use the Word "Would" Like This: "\$160 would be a good price."}\\
&&\textit{...}\\
& S: &\textbf{I always took great care of the truck.} I think \$10,750 \textbf{would} be a reasonable price.\\
\textit{29.} & B: &Deal.\\
\textit{30.} & S: &Great\\
 \bottomrule
\end{tabularx}
\caption{Examples of collected negotiation dialogs.} 
\label{table:case_study}
\end{table*}

\end{document}